\documentclass[a4paper]{article}

\usepackage{INTERSPEECH2019}
\usepackage[colorlinks=true,linkcolor=black,anchorcolor=black,citecolor=black,filecolor=black,menucolor=black,runcolor=black,urlcolor=black]{hyperref}
\usepackage{adjustbox}

\title{Detecting Spoofing Attacks Using VGG and SincNet: \\
BUT-Omilia Submission to ASVspoof 2019 Challenge}

\name{ Hossein Zeinali$^1$, Themos Stafylakis$^2$, Georgia Athanasopoulou$^2$,  Johan Rohdin$^1$ \\
	Ioannis Gkinis$^2$, Luk\'a\v{s} Burget$^1$, and Jan ``Honza'' \v{C}ernock\'{y}$^{1}$ }
\address{$^1$Brno University of Technology, Faculty of IT, IT4I Centre of Excellence, Czechia \\
	$^2$Omilia - Conversational Intelligence, Athens, Greece
}
\email{\{zeinali,rohdin,burget,cernocky\}@fit.vutbr.cz \{tstafylakis,gathanasopoulou,igkinis\}@omilia.com}

\begin{document}

\maketitle

\begin{abstract}
In this paper, we present the system description of the joint efforts of Brno University of Technology (BUT) and Omilia -- Conversational Intelligence for the ASVSpoof2019 Spoofing and Countermeasures Challenge. The primary submission for Physical access (PA) is a fusion of two VGG networks, trained on single and two-channels features. For Logical access (LA), our primary system is a fusion of VGG and the recently introduced SincNet architecture. The results on PA show that the proposed networks yield very competitive performance in all conditions and achieved 86\:\% relative improvement compared to the official baseline. On the other hand, the results on LA showed that although the proposed architecture and training strategy performs very well on certain spoofing attacks, it fails to generalize to certain attacks that are unseen during training.
\end{abstract}

\section{Introduction}
\label{sec:intro}

To facilitate better and safer customer support in e.g. banking and call centers, there is a growing demand for convenient and robust automatic authentication systems. Automatic speaker verification (ASV) a.k.a. voice biometrics is arguably the most natural and least intrusive authentication method in such applications.
Unfortunately, ASV systems are vulnerable to synthetic speech, created by text-to-speech (TTS) and voice conversion (VC) methods, and to replay/presentation attacks~\cite{wu2015spoofing}. The attempts to deceive an ASV system by such methods are known as {\it ASV spoofing} attacks. While research in ASV has been ongoing for several decades, it is only in the recent years that the research community has started to tackle spoofing attacks systematically, through a series of ASV spoofing and countermeasures challenges~\cite{wu2015asvspoof, kinnunen2017asvspoof}.

Spoofing attacks to ASV systems can be categorized into 4 types~\cite{wu2015spoofing}. The first one is impersonation which can be rejected by an accurate ASV system~\cite{hautamaki2013vectors}. The second and third types are TTS and VC which were tackled in the ASVspoof 2015 challenge~\cite{wu2015asvspoof} and several methods have been proposed to detect them~\cite{alam2015development, chen2015robust, novoselov2016stc, patel2015combining}. The last type of attacks is replay attack with pre-recorded audio and it is considered to be the most difficult attack to detect~\cite{wu2015spoofing}. Possible ways to tackle this problem are (a) anti-spoofing techniques based on detecting typical distortions in recorded and replayed audio~\cite{kinnunen2017asvspoof, lavrentyeva2017audio}, (b) using audio fingerprinting~\cite{gonzalez2018audio} to detect a replay of an enrollment utterance, and (c) using liveness detection and phrase verification~\cite{zeinali2018spoken} in text-dependent speaker verification.
 
This paper presents the collaborative efforts of BUT and Omilia to introduce novel countermeasures for the last three attack types, as part of the 2019 automatic speaker verification (ASV) anti-spoofing challenge.
All our systems are based on deep neural network (DNN) architectures, trained to discriminate between bonafide and synthetic or replayed speech and are employed as end-to-end classifiers, i.e. without any external backend. 
The physical access (PA) system is a fusion of two VGG~\cite{simonyan2014very} networks using different features, while the logical access (LA) system is a fusion of one VGG network and two SincNet networks~\cite{ravanelli2018speaker}.


\section{Physical access}

\subsection{Features and preprocessing}

For this challenge we explore several features such as Mel-filter bank, MFCC, constant Q-transform (CQT)~\cite{schorkhuber2010constant}, CQCC~\cite{todisco2016new}, and power spectrogram. Among the explored features, power spectrogram yields superior performance, followed by CQT features. Accordingly, we use these two features in most of our experiments. In particular, the submitted systems use either the power spectrograms as a single input channel, or both the power spectrograms and the CQT features fed as two different input channels. As a feature preprocessing, both CQT and power spectrogram are first transferred to log domain and then subjected to mean and variance normalization (MVN) before being fed to the network.

\subsection{Example and minibatch generation for network training}

The procedure for generating training examples and minibatches can greatly affect the performance of neural networks in audio processing. Therefore, we experimented with several different strategies for this. For example generation, we first concatenate all features of the same class (same attack id) and speaker. We then split the concatenated features into small segments of the same size. Initially we used four second segments but after doing several experiments, we found that networks trained on smaller segments performed better than those trained on large segments, mainly because they overfit less to the training data. The size of the examples used to train the submitted systems is one second (i.e. 100 frames).

For minibatch generation we experimented with different strategies for distributing the examples into minibatches. We found that the best strategy is to only use examples from a single speaker within each minibatch (a few minibatches may contain examples from more speakers in order to use all training data). Each minibatch has 128 examples. After each epoch, we randomise the examples and generate the minibatches again for better generalization.

\subsection{Training and development data}

For training the networks, the official training set of the challenge was used. This set contains audio samples from 20 speakers. One of the speakers was randomly selected for network training validation set which is roughly 5\:\% of the training data.

The development set is also the official challenge's development set. This set which contains 20 speakers, was only used for evaluating networks and comparing different methods and training strategies.

\subsection{Networks and training strategies}

For this challenge, two different topologies were used for Physical access. The first one is a modified version of a VGG network~\cite{simonyan2014very} which has shown good performance in Audio Tagging and Audio Scene Classification~\cite{Dorfer2018, Zeinali2018}. The second network is a modified version of a Light CNN (LCCN)~\cite{wu2018light} which had the best performance for ASVSpoof2017 challenge~\cite{lavrentyeva2017audio}. We have used a modified version of both networks for acoustic scene classification challenge 2019~\cite{zeinali2019dcase}. In the following two sections, both networks will be explained in more detail.

\vspace{-0.5mm}
\subsubsection{VGG-like network}

The VGG network comprises several convolutional and pooling layers followed by a statistics pooling and several dense layers which perform classification. Table~\ref{tab:vgg} provides a detailed description of the proposed VGG architecture. There are 6 convolutional blocks in the model, each containing 2 convolutional layers and one max-pooling. Each max-pooling layer reduce the size of frequency axis to half while only one of them reduces the temporal resolution. After the convolutional layers, there is a mean pooling layer which operates only on the time axis and calculates the mean over time. After this layer, there is a flatten layer which simply concatenates the 4 remaining frequency channels. Finally there are 3 dense layers which perform the classification task.

\begin{table}[t]
  \renewcommand{\arraystretch}{0.8}
	\centering
	\caption{\label{tab:vgg}The proposed VGG architecture. Conv2D: two  dimensional convolutional layer. MeanPooling: a layer which calculate the mean in time axis and reduce the shape (remove the time axis). Dense: fully connected dense layer.}
	\vspace{-2mm}
	\setlength\tabcolsep{5pt}
	\begin{tabular}{l l l r}
		\toprule
		\toprule
		\textbf{Layer name}   & \textbf{Filter}   & \textbf{Output}                   & \textbf{\#Params} \\
		\midrule
		Input                 & --                & 256 $\times$ 100 $\times$ 2       & -- \\
		Conv2D-1-1            & 3 $\times$ 3      & 256 $\times$ 100 $\times$ 32      & 608 \\
		Conv2D-1-1            & 3 $\times$ 3      & 256 $\times$ 100 $\times$ 32      & 9.2K \\
		MaxPooling-1          & 2 $\times$ 1      & 128 $\times$ 100 $\times$ 32      & -- \\
		\midrule
		Conv2D-2-1            & 3 $\times$ 3      & 128 $\times$ 100 $\times$ 64      & 18.5K \\
		Conv2D-2-2            & 3 $\times$ 3      & 128 $\times$ 100 $\times$ 64      & 37K \\
		MaxPooling-2          & 2 $\times$ 1      & 64  $\times$ 100 $\times$ 64      & --  \\
		\midrule
		Conv2D-3-1            & 3 $\times$ 3      & 64 $\times$ 100 $\times$ 128      & 74K  \\
		Conv2D-3-2            & 3 $\times$ 3      & 64 $\times$ 100 $\times$ 128     & 148K \\
		MaxPooling-3          & 2 $\times$ 2      & 32 $\times$ 50  $\times$ 128     & --   \\
		\midrule
		Conv2D-4-1            & 3 $\times$ 3      & 32 $\times$ 50 $\times$ 256     & 295K  \\
		Conv2D-4-2            & 3 $\times$ 3      & 32 $\times$ 50 $\times$ 256     & 590K \\
		MaxPooling-4          & 2 $\times$ 1      & 16 $\times$ 50 $\times$ 256     & --   \\
		\midrule
		Conv2D-5-1            & 3 $\times$ 3      & 16 $\times$ 50 $\times$ 256     & 590K  \\
		Conv2D-5-2            & 3 $\times$ 3      & 16 $\times$ 50 $\times$ 256     & 590K \\
		MaxPooling-5          & 2 $\times$ 1      & 8  $\times$ 50 $\times$ 256     & --   \\
		\midrule
		Conv2D-6-1            & 3 $\times$ 3      & 8 $\times$ 50 $\times$ 256     & 590K  \\
		Conv2D-6-2            & 3 $\times$ 3      & 8 $\times$ 50 $\times$ 256     & 590K \\
		MaxPooling-6          & 2 $\times$ 1      & 4 $\times$ 50 $\times$ 256     & --   \\
		\midrule
		MeanPooling           & --                & 4 $\times$ 256                 & --  \\
		Flatten               & --                & 1024                           & --  \\
		\midrule
		Dense1                & --                & 512                            & 525K  \\
		Dense2                & --                & 512                            & 263K  \\
		Dense3 (softmax)       & --                & 2                              & 1K  \\
		\midrule
		Total                 & --                & --                             & 4321K  \\
		\bottomrule
		\bottomrule
	\end{tabular}
	\vspace{-2mm}
\end{table}

\vspace{-0.5mm}
\subsubsection{Light CNN (LCNN)}

Table~\ref{tab:lcnn} shows the used LCNN topology for this challenge. This network is a combination of convolutional and max-pooling layers and uses Max-Feature-Map (MFM) as non-linearity. MFM is a layer which simply reduce the number of output channels to the half by taking the maximum of two consecutive channels (or any other combination of two channels). The rest of this network (statistics and classification parts) is identical to the proposed VGG network.

\begin{table}[t]
  \renewcommand{\arraystretch}{0.8}
	\centering
	\caption{\label{tab:lcnn} The proposed LCNN architecture. MFM: Max-Feature-Map.}
	\vspace{-2mm}
	\setlength\tabcolsep{5pt}
	\begin{tabular}{l l l r}
		\toprule
		\toprule
		\textbf{Layer name}   & \textbf{Filter}   & \textbf{Output}                   & \textbf{\#Params} \\
		\midrule
		Input                 & --                & 256 $\times$ 100 $\times$ 2       & -- \\
		Conv2D-1-1            & 5 $\times$ 5      & 256 $\times$ 100 $\times$ 32      & 1K \\
		MFM-1-1               & --                & 256 $\times$ 100 $\times$ 16      & -- \\
		MaxPooling-1          & 2 $\times$ 1      & 128 $\times$ 100 $\times$ 16      & -- \\
		\midrule
		Conv2D-2-1            & 1 $\times$ 1      & 128 $\times$ 100 $\times$ 32      & 544 \\
		MFM-2-1               & --                & 128 $\times$ 100 $\times$ 16      & -- \\
		Conv2D-2-2            & 3 $\times$ 3      & 128 $\times$ 100 $\times$ 64      & 10K \\
		MFM-2-2               & --                & 128 $\times$ 100 $\times$ 32      & -- \\
		MaxPooling-2          & 2 $\times$ 1      & 64  $\times$ 100 $\times$ 32      & --  \\
		\midrule
		Conv2D-3-1            & 1 $\times$ 1      & 64 $\times$ 100 $\times$ 64      & 74K  \\
		MFM-3-1               & --                & 64 $\times$ 100 $\times$ 32      & -- \\
		Conv2D-3-2            & 3 $\times$ 3      & 64 $\times$ 100 $\times$ 128     & 37K \\
		MFM-3-2               & --                & 64 $\times$ 100 $\times$ 64      & -- \\
		MaxPooling-3          & 2 $\times$ 2      & 32 $\times$ 50  $\times$ 64      & --   \\
		\midrule
		Conv2D-4-1            & 1 $\times$ 1      & 32 $\times$ 50 $\times$ 128     & 8K  \\
		MFM-4-1               & --                & 32 $\times$ 50 $\times$ 64      & -- \\
		Conv2D-4-2            & 3 $\times$ 3      & 32 $\times$ 50 $\times$ 256     & 148K \\
		MFM-4-2               & --                & 32 $\times$ 50 $\times$ 128     & -- \\
		MaxPooling-4          & 2 $\times$ 1      & 16 $\times$ 50 $\times$ 128     & --   \\
		\midrule
		Conv2D-5-1            & 1 $\times$ 1      & 16 $\times$ 50 $\times$ 256     & 33K  \\
		MFM-5-1               & --                & 16 $\times$ 50 $\times$ 128     & -- \\
		Conv2D-5-2            & 3 $\times$ 3      & 16 $\times$ 50 $\times$ 512     & 590K \\
		MFM-5-2               & --                & 16 $\times$ 50 $\times$ 256     & -- \\
		MaxPooling-5          & 2 $\times$ 1      & 8  $\times$ 50 $\times$ 256     & --   \\
		\midrule
		Conv2D-6-1            & 1 $\times$ 1      & 8 $\times$ 50 $\times$ 512     & 132K  \\
		MFM-6-1               & --                & 8 $\times$ 50 $\times$ 256     & -- \\
		Conv2D-6-2            & 3 $\times$ 3      & 8 $\times$ 50 $\times$ 512     & 1180K \\
		MFM-6-2               & --                & 8 $\times$ 50 $\times$ 256     & -- \\
		MaxPooling-6          & 2 $\times$ 1      & 4 $\times$ 50 $\times$ 256     & --   \\
		\midrule
		MeanPooling           & --                & 4 $\times$ 256                 & --  \\
		Flatten               & --                & 1024                           & --  \\
		\midrule
		Dense1                & --                & 512                            & 525K  \\
		Dense2                & --                & 512                            & 263K  \\
		Dense (softmax)       & --                & 2                              & 1K  \\
		\midrule
		Total                 & --                & --                             & 2930K  \\
		\bottomrule
		\bottomrule
	\end{tabular}
	\vspace{-3mm}
\end{table}

\subsection{Fusion and submitted systems}

Since the evaluation protocol does not allow us to estimate fusion parameters on the development set, we choose to use a simple average with equal weight for our best systems. Our submissions are the following:
\begin{itemize}
	\item {\bf Primary:} Fusion of two VGG networks. The first one is trained using two-channels features while the second one is fed with single channel log-power spectrogram.
	\item {\bf Single best:} Our single best system for this part is the VGG network with two-channels features.
	\item {\bf Contrastive 1:} This system is a VGG network with single channel log-power spectrogram features.
	\item {\bf Contrastive 2:} The second contrastive system is LCNN network again with single channel log-power spectrogram as features.
\end{itemize}

\section{Logical access}

\subsection{Logical access using SincNet}

SincNet is a novel end-to-end neural network architecture, which receives raw waveforms as input rather than handcrafted features such as spectrograms or CQCCs~\cite{ravanelli2018speaker}. Contrary to other end-to-end approaches, SincNet constrains the first 1D convolutional layer to parametrized Sinc functions, encouraging it to discover more meaningful (band-pass) filters. This architecture offers a very efficient way to derive a customized filter bank that is specifically tuned for the desired application. The filters are initialized using the Mel-frequency filter bank and their low and high cutoff frequencies are adapted with standard back-propagation as any other layer. SincNet is originally designed for speech and speaker recognition tasks, and we believe it is a good fit for the problem at hand, since certain artifacts created by TTS and VC systems should be more easily detectable in the waveform domain.

\subsubsection{SincNet architecture}

The first block consists of three convolutional layers. The first layer performs Sinc-based convolutions, using 80 filters of length L=251 samples. The remaining two layers using 60 filters of length 5. Next, three fully-connected layers composed of 2048 neurons and normalized with batch normalization were applied. All hidden layers use leaky-ReLU non-linearities. Frame-level binary classification is performed by applying a softmax classifier and cross-entropy criterion. We use high dropout rates in all layers in one of our networks, in order to improve its generalizability to unseen speakers and spoofing attacks~\cite{ravanelli2018speaker}. Our implementation is based on the open-source PyTorch code provided by the authors \footnote{\url{https://github.com/mravanelli/SincNet}}.

\subsubsection{Training and evaluating SincNet}

SincNet is trained by randomly sampling 200\:ms chunks from each utterance, which are fed into the SincNet architecture. Mean and variance normalization and energy-based voice activity detector are applied in an utterance-level fashion. As in the original SincNet we use RMSprop as optimizer, while we train it with only 5 epochs, each comprising 1000 minibatches of size 256. In the first epoch, we use a small learning rate, which we increase and decrease again for the last epoch (namely $10^{-5},10^{-4},10^{-3}$ and $10^{-4}$). The small learning rate in the first epoch is chosen in order to preserve the mel-frequency based initialization of the Sinc functions. This learning rate approach results to a steep decrease in the loss from the fourth epoch. Moreover, during training we ensure that each minibatch used for back-propagation is balanced, such that for every bonafide sample we randomly select a spoof sample from the same speaker, resulting in 128 bonafide samples and 128 spoof samples for every minibatch.

During evaluation, utterance-level LLRs are derived by averaging the corresponding frame-level LLRs, as estimated by the logarithmic softmax layer.

\subsubsection{Cross-validation over presentation attacks}

In order to assess the generalizability of the network to novel attacks, we first trained the network on a subset of attacks and evaluated it on the remaining ones. By using this cross-validation scheme, the EER attained on unseen attacks was always below 0.2\% EER, underlying the good generalization capacity of the network, at least between those attacks included in the training and development sets. Finally, we trained the model on the whole training set using the best training strategy defined by the cross-validation and we obtained 0.0\:\% EER (i.e. no errors) on the full development set.  

\subsection{Logical access using VGG}

For the Logical access we explored the two VGG architectures that were the best for Physical access, i.e. the architecture described in Table \ref{tab:vgg} with either log-power spectrum as a single input channel, or with log-power-spectrum and CQT as two input channels. Using only the log-power spectrum was substantially better than using both features.

It is worth noting that we experimented with the SincNet architecture on presentation attacks (i.e. PA), however its performance was inferior to that of VGG.

\subsection{Fusion and submitted systems}

As in physical access we have 4 systems and again we fuse them using simple averaging.
\begin{itemize}
	\item {\bf Primary:} Our primary system is fusion of a VGG network with single channel log-power spectrogram features and 2 SincNets which differ in the dropout rate. 
	\item {\bf Single best:} SincNet with the standard dropout rates.
	\item {\bf Contrastive 1:} Fusion of two VGG network which were trained using two channel and single channel features like Physical access.
	\item {\bf Contrastive 2:} SincNet with high dropout rates.
\end{itemize}

\section{Experimental results}

In this section, we report the official results evaluated by the challenge organizers, based on the scores we submitted. 

\begin{table*}[th]
	\renewcommand{\arraystretch}{1}
	\caption{\label{tbl.PA_detailed_results} Physical access detailed results based on min-tDCF for different conditions. The first section shows the baseline results and the second section shows the primary and single best results of the best-performing systems, both from team T28.}
	\vspace{-2mm}
	\centerline
	{
		\setlength\tabcolsep{3pt}
		\begin{adjustbox}{width=1.0\textwidth}
			\begin{tabular}{ l c c c c c c c c c c c c c c c c c c c c c c c c c c c c }
				\toprule
				\midrule
				& \multicolumn{9}{c}{Development set} & & \multicolumn{9}{c}{Evaluation set} \\
				\cmidrule{2-10} \cmidrule{12-20}
				\textbf{System}	    & \textbf{AA} & \textbf{AB} & \textbf{AC} & \textbf{BA} & \textbf{BB} &  \textbf{BC} & \textbf{CA} & \textbf{CB} & \textbf{CC} & & \textbf{AA} & \textbf{AB} & \textbf{AC} & \textbf{BA} & \textbf{BB} &  \textbf{BC} & \textbf{CA} & \textbf{CB} & \textbf{CC} \\
				\midrule
				CQCC-GMM			& 0.4928 & 0.0539 & 0.0213 & 0.3999 & 0.0360 & 0.0197 & 0.4338 & 0.0414 & 0.0149 & & 0.4975 & 0.1751 & 0.0529 & 0.4658 & 0.1483 & 0.0433 & 0.5025 & 0.1360 & 0.0461 \\
				\midrule
				T28 Primary			& 0.0132 & 0.0030 & 0.0009 & 0.0073 & 0.0017 & 0.0009 & 0.0065 & 0.0023 & 0.0008 & & 0.0190 & 0.0079 & 0.0034 & 0.0113 & 0.0083 & 0.0022 & 0.0127 & 0.0075 & 0.0024 \\
				T28 Single			& 0.0185 & 0.0044 & 0.0013 & 0.0146 & 0.0043 & 0.0014 & 0.0146 & 0.0081 & 0.0024 & & 0.0251 & 0.0107 & 0.0055 & 0.0152 & 0.0114 & 0.0058 & 0.0183 & 0.0111 & 0.0063 \\
				\midrule
				Primary				& 0.0389 & 0.0062 & 0.0039 & 0.0243 & 0.0049 & 0.0048 & 0.0233 & 0.0073 & 0.0028 & & 0.0776 & 0.0217 & 0.0091 & 0.0586 & 0.0223 & 0.0088 & 0.0557 & 0.0256 & 0.0110 \\
				Single best			& 0.0611 & 0.0046 & 0.0040 & 0.0404 & 0.0052 & 0.0053 & 0.0402 & 0.0085 & 0.0039 & & 0.1061 & 0.0267 & 0.0117 & 0.0901 & 0.0277 & 0.0115 & 0.0843 & 0.0330 & 0.0128 \\
				Contrastive1		& 0.0523 & 0.0245 & 0.0151 & 0.0256 & 0.0156 & 0.0130 & 0.0280 & 0.0229 & 0.0135 & & 0.0695 & 0.0383 & 0.0148 & 0.0493 & 0.0383 & 0.0141 & 0.0437 & 0.0394 & 0.0192 \\
				Contrastive2		& 0.0726 & 0.0323 & 0.0170 & 0.0562 & 0.0283 & 0.0153 & 0.0633 & 0.0353 & 0.0167 & & 0.0969 & 0.0547 & 0.0187 & 0.0843 & 0.0519 & 0.0193 & 0.0842 & 0.0532 & 0.0229 \\
				\midrule
				\bottomrule
			\end{tabular}
		\end{adjustbox}
		\vspace{-2mm}
	}
\end{table*}

\begin{table*}[th]
	\renewcommand{\arraystretch}{1.0}
	\caption{\label{tbl.LA_detailed_results} Logical access detailed results based on min-tDCF for different conditions. The first section shows the baseline results and the second section shows the primary system results of the best performing team (T05) as well as the overall best single system results (team T45). The bold numbers show conditions where our single system performs better or the same as the best single system.}
	\vspace{-2mm}
	\centerline
	{
		\setlength\tabcolsep{3pt}
		\begin{adjustbox}{width=1.0\textwidth}
		\begin{tabular}{ l c c c c c c c c c c c c c c c c c c c c c c c c c c c c  }
			\toprule
			\midrule
			& \multicolumn{6}{c}{Development set} & & \multicolumn{13}{c}{Evaluation set} \\
			\cmidrule{2-7} \cmidrule{9-21}
			\textbf{System}	    & \textbf{A01} & \textbf{A02} & \textbf{A03} & \textbf{A04} & \textbf{A05} &  \textbf{A06} & & \textbf{A07} & \textbf{A08} & \textbf{A09} & \textbf{A10} & \textbf{A11} & \textbf{A12} & \textbf{A13} & \textbf{A14} & \textbf{A15} & \textbf{A16} & \textbf{A17} & \textbf{A18} & \textbf{A19} \\
			\midrule
			CQCC-GMM			& 0.0000 & 0.0000 & 0.0020 & 0.0000 & 0.0261 & 0.0011 & & 0.0000 & 0.0007 & 0.0060 & 0.4149 & 0.0020 & 0.1160 & 0.6729 & 0.2629 & 0.0344 & 0.0000 & 0.9820 & 0.2818 & 0.0014 \\
			\midrule
			T05 Primary			& 0.0000 & 0.0000 & 0.0000 & 0.0000 & 0.0000 & 0.0000 & & 0.0009 & 0.0014 & 0.0000 & 0.0077 & 0.0055 & 0.0045 & 0.0028 & 0.0035 & 0.0050 & 0.0015 & 0.0341 & 0.0276 & 0.0020 \\
			T45 Single			& 0.0027 & 0.0000 & 0.0000 & 0.0036 & 0.0068 & 0.0085 & & 0.0034 & 0.0308 & 0.0000 & 0.0130 & 0.0017 & 0.0058 & 0.0034 & 0.0042 & 0.0065 & 0.0071 & 0.9833 & 0.1171 & 0.0895 \\
			\midrule
			Primary				& 0.0000 & 0.0000 & 0.0000 & 0.0000 & 0.0000 & 0.0000 & & 0.0000 & 0.0000 & 0.0029 & 0.5672 & 0.0425 & 0.0425 & 0.1098 & 0.0005 & 0.5525 & 0.0000 & 0.3775 & 0.6473 & 0.0000 \\
			Single best			& \textbf{0.0000} & \textbf{0.0000} & \textbf{0.0000} & \textbf{0.0000} & \textbf{0.0000} & \textbf{0.0000} & & \textbf{0.0002} & \textbf{0.0004} & 0.1393 & 0.9423 & 0.0426 & 1.0000 & 0.3693 & \textbf{0.0000} & 1.0000 & \textbf{0.0004} & \textbf{0.4764} & 0.6731 & \textbf{0.0000} \\
			Contrastive1		& 0.0000 & 0.0000 & 0.0000 & 0.0000 & 0.0003 & 0.0000 & & 0.0654 & 0.2004 & 0.1663 & 0.5031 & 0.0002 & 0.9297 & 0.8583 & 0.0000 & 0.0002 & 0.0007 & 0.0263 & 0.5749 & 0.3217 \\
			Contrastive2		& 0.0000 & 0.0000 & 0.0000 & 0.0010 & 0.0000 & 0.0000 & & 0.0017 & 0.0026 & 0.1505 & 0.9992 & 0.0253 & 1.0000 & 0.4737 & 0.0000 & 1.0000 & 0.0022 & 0.4131 & 0.9420 & 0.0009 \\
			\midrule
			\bottomrule
		\end{tabular}
	\end{adjustbox}
		\vspace{-4mm}
	}
\end{table*}

\subsection{Results on Physical Access}

Table~\ref{tbl.PA_results} reports results attained by different submissions for physical access. The first row of the table provides the results for the organizers' baseline which is a GMM based method with CQCC features. The results on the evaluation set attained by our submitted systems demonstrate their capacity in generalizing very well to new PA configurations. By comparing the single best and contrastive1 systems it is evident that the single channel features perform considerably better on the evaluation set (has better generalization).

A more analytic report can be found in Table \ref{tbl.PA_detailed_results}. The first letter in attack ID shows the environment definition. From A to C, room size, room reverberation time and talker-to-ASV distance are increased and so, detection of A is more difficult than C. The second letter of attack ID shows attack definition. From A to C, attacker-to-talker distance is increased while replay device quality is decreased. Again, A is more difficult than C. It is clear that the trends of the results are in line with expectations in most cases (i.e. AA has the worst results and CC has the best.)

\begin{table}[t]
	\renewcommand{\arraystretch}{0.9}
	\caption{\label{tbl.PA_results} Physical access results of different submissions}
	\vspace{-2mm}
	\centerline
	{
		\setlength\tabcolsep{2.6pt}
		\begin{tabular}{ l c c c c c }
			\toprule
			\midrule
								& \multicolumn{2}{c}{Development set} & & \multicolumn{2}{c}{Evaluation set} \\
								\cmidrule{2-3} \cmidrule{5-6}
			\textbf{System}	    & \textbf{EER[\%]} & \textbf{min-tDCF} & & \textbf{EER [\%]} & \textbf{min-tDCF} \\
			\midrule
			CQCC-GMM			& 9.87 & 0.1953 & & 11.04 & 0.2454 \\
			\midrule
			Primary				& 0.66 & 0.0170 & & 1.51  & 0.0372 \\
			Single best			& 1.02 & 0.0254 & & 2.11  & 0.0527 \\
			Contrastive1		& 1.07 & 0.0253 & & 1.49  & 0.0401 \\
			Contrastive2		& 1.59 & 0.0401 & & 2.31  & 0.0591 \\
			\midrule
			\bottomrule
		\end{tabular}
		\vspace{-3mm}
	}
\end{table}

\subsection{Results on Logical Access}

We present here the results we attained on the evaluation test. In Table~\ref{tbl.LA_results} we report the results on the two sets. Clearly, although our systems performed exceptionally well on the development set, failed to generalize well to certain logical attacks unseen in training. 

The LA detailed results are reported in Table \ref{tbl.LA_detailed_results} based on different waveform generation methods include: {\it neural waveform} (A01, A08, A10, A12, A15), {\it vocoder} (A02, A03, A07, A09, A14, A18), {\it waveform filtering} (A05, A13, A17), {\it spectral filtering} (A06, A19) and {\it waveform concatenation} (A04, A13, A16).
From the table, we observe that the attacks which degraded the performance the most were A10, A12, and A15, which were all based on neural waveform TTS systems. It is interesting to note that for these attacks, the EER attained by SincNet was above 50\:\% (not reported here) while it performs better than or same as the overall best single system in 12 conditions. The conclusion is that the cross-validation method we performed was insufficient to prevent the network from overfitting and some more analysis will be needed to figure out why the SincNet totally failed for some waveform generation methods.

\section{Conclusions}

In this paper we presented the joint submission of BUT and Omilia for the ASVspoof 2019. For PA, we followed the VGG architecture and obtained very competitive results in both development and evaluation sets, by fusing only two networks. For LA, we fused a VGG architecture with the recently proposed SincNet. The rationale for employing the latter was its ability to jointly optimize the networks and the feature extractor, which was shown to be very effective for speech and speaker recognition. Despite our efforts to prevent overfitting (mainly via attack-level cross validation in training and development), the results on LA showed the difficulty of the SincNet in generalizing to certain attacks which were significantly different to those in the training. We conclude that more research is required in order to make full use of end-to-end anti-spoofing architectures such as SincNet in cases of large mismatch between training and evaluation attacks.

\begin{table}[t]
	\renewcommand{\arraystretch}{0.9}
	\caption{\label{tbl.LA_results} Logical access results of different submissions}
	\vspace{-2mm}
	\centerline
	{
		\setlength\tabcolsep{2.6pt}
		\begin{tabular}{ l c c c c c }
			\toprule
			\midrule
			& \multicolumn{2}{c}{Development set} & & \multicolumn{2}{c}{Evaluation set} \\
			\cmidrule{2-3} \cmidrule{5-6}
			\textbf{System}	    & \textbf{EER[\%]} & \textbf{min-tDCF} & & \textbf{EER [\%]} & \textbf{min-tDCF} \\
			\midrule
			CQCC-GMM			& 0.43 & 0.0123 & & 9.57  & 0.2366 \\
			\midrule
			Primary				& 0.00 & 0.0000 & & 8.01  & 0.2080 \\
			Single best			& 0.00 & 0.0000 & & 20.11 & 0.3563 \\
			Contrastive1		& 0.00 & 0.0000 & & 10.52 & 0.2790 \\
			Contrastive2		& 0.03 & 0.0003 & & 22.99 & 0.3811 \\
			\midrule
			\bottomrule
		\end{tabular}
	}
\end{table}

\section{Acknowledgment}
The work was supported by Czech Ministry of Education, Youth and Sports from Project No. CZ.02.2.69/0.0/0.0/16\_027/0008371, the National Programme of Sustainability (NPU II) project IT4Innovations excellence in science - LQ1602, the Marie Sklodowska-Curie cofinanced by the South Moravian Region under grant agreement No. 665860, and by Czech Ministry of Interior project No. VI20152020025 "DRAPAK".

\bibliographystyle{IEEEtran}

\bibliography{mybib}

\end{document}